%% file: coling2020.tex
\documentclass[11pt]{article}
\usepackage{coling2020}
\usepackage{kotex}
\usepackage{mathtools}
\usepackage{adjustbox}
\usepackage{times}
\usepackage{url}
\usepackage{latexsym}
\usepackage{amsmath}
\usepackage{array}
\usepackage{xcolor}
\usepackage{multirow}
\usepackage{booktabs}
\usepackage{graphicx}
\usepackage{subcaption}
\usepackage{float}
\usepackage{amssymb}

\newcolumntype{L}{>{\arraybackslash}m{20cm}}

\colingfinalcopy 

\title{Reference and Document Aware Semantic Evaluation Methods for Korean Language Summarization}

\author{Dongyub Lee\textsuperscript{1}, Myeongcheol Shin\textsuperscript{2}, Taesun Whang\textsuperscript{3}, Seungwoo Cho\textsuperscript{2}, \\ \textbf{Byeongil Ko}\textsuperscript{2}, \textbf{Daniel Lee}\textsuperscript{2}, \textbf{Eunggyun Kim}\textsuperscript{2}, \textbf{Jaechoon Jo\textsuperscript{4}\thanks{\hspace{0.1cm} corresponding author}}  \\
  \textsuperscript{1}Kakao Corp. \\
  {\small\tt jude.lee@kakaocorp.com} \\
  \textsuperscript{2}Kakao Enterprise Corp.\\
  {\small\tt \{index.sh,john.w,kobi.k,daniel.e,json.ng\}@kakaoenterprise.com} \\
  \textsuperscript{3}Korea University, South Korea \\
  {\small\tt \{taesunwhang\}@korea.ac.kr} \\
 \textsuperscript{4}Hanshin University, South Korea \\
  {\small\tt \{jaechoon\}@hs.ac.kr} \\  
}

\date{}

\begin{document}
\maketitle
\begin{abstract}
  Text summarization refers to the process that generates a shorter form of text from the source document preserving salient information. Many existing works for text summarization are generally evaluated by using recall-oriented understudy for gisting evaluation (ROUGE) scores. However, as ROUGE scores are computed based on n-gram overlap, they do not reflect semantic meaning correspondences between generated and reference summaries. Because Korean is an agglutinative language that combines various morphemes into a word that express several meanings, ROUGE is not suitable for Korean summarization. In this paper, we propose evaluation metrics that reflect semantic meanings of a reference summary and the original document, \textit{~\textbf{R}eference and ~\textbf{D}ocument ~\textbf{A}ware ~\textbf{S}emantic ~\textbf{S}core (RDASS)}. We then propose a method for improving the correlation of the metrics with human judgment. Evaluation results show that the correlation with human judgment is significantly higher for our evaluation metrics than for ROUGE scores.

\end{abstract}

\section{Introduction}
\label{intro}
The task of text summarization is to generate a reference summary that conveys all the salient information of an original document. There are two strategies for this type of summarization (i.e., extractive and abstractive summarization). With the extractive approach, the most noticeable key sentences are extracted from the source and compiled into a reference~\cite{zhong2019searching,wang2019self,xiao2019extractive}. The second approach is abstractive, with which a paraphrased summary is generated from the source~\cite{zhang2018abstractiveness,guo2018soft,wenbo2019concept}. The generated summary may not contain the same words that appear in the source document. Therefore, measuring factual alignment between the generated summary and source document is important~\cite{kryscinski2019neural}.

Most summarization models are evaluated using recall-oriented understudy for gisting evaluation (ROUGE)~\cite{lin-2004-rouge}, which measures n-gram overlaps between generated and reference summaries. ROUGE has proven to have a high correlation with manual evaluation methods, such as pyramid~\cite{nenkova2007pyramid} and TAC AESOP~\cite{owczarzak2011overview}. However, Louis~\shortcite{louis2013automatically} showed that the correlation significantly decreased when only one reference summary was provided. Additionally, considering the process by which a person manually summarizes a document, ROUGE is limited, because it does not reflect semantic meanings between generated and reference summaries. For example, when a person summarizes a document, they tend to use words that are implicit while not always using the explicit words from the original document. As the ROUGE score is computed based on an n-gram overlap, the score can be low even if two words have the same semantic meaning. Table~\ref{limit_table} shows an example of the ROUGE limitation when applied to a Korean summarization. This tendency is particularly prevalent in Korean, which is an agglutinative language that combines various morphemes into a word to express several meanings and grammatical functions, unlike English. In this process, complex morphological variations can occur. Therefore, leveraging ROUGE scores produces inaccurate results. 

\begin{table}[h]
\begin{adjustbox}{width=1.0\textwidth}
    \centering
    \begin{tabular}{|l|l|}
    \hline
    \multicolumn{2}{|l|}{\textbf{Article}}            \\ \hline
    \multicolumn{2}{|L|}{ `슬기로운 의사생활'이 또다시 최고 시청률을 경신하며 고공행진을 이어갔다. 26일 방송된 tvN 2020 목요 스페셜 `슬기로운 의사생활' 3회는 케이블, IPTV, 위성을 통합한 유료플랫폼에서 가구 평균 8.6\%, 최고 10\%의 시청률을 기록했다. 3주 연속 시청률 상승세다. \newline The tv program ``sage doctor life'' set an all time record for its highest viewer ratings. On the 26th, TVN 2020 Thursday Special ``Sage Doctor Life,'' aired on the 26th, recorded an average household rating of 8.6\% and a maximum of 10\% on a paid platform incorporating cable, IPTV, and satellite. The ratings have been rising for three consecutive weeks.}              \\ \hline
    \multicolumn{2}{|l|}{\textbf{Reference Summary}}            \\ \hline
    \multicolumn{2}{|L|}{ `슬기로운 의사생활' \textcolor{blue}{최고} 시청률 10\% 돌파… 3회 연속 \textcolor{blue}{상승}세 \newline The tv program ``Sage doctor life'' breaks its all time high 10\% viewer ratings for 3 consecutive episodes.}              \\ \hline
    
    \multicolumn{2}{|l|}{\textbf{Wrong Candidate}}            \\ \hline
    \multicolumn{2}{|L|}{ `슬기로운 의사생활' \textcolor{red}{최저} 시청률 10\% 돌파... 3회 연속 \textcolor{red}{하락}세 \newline The tv program ``Sage doctor life'' reached the lowest viewing rate of 10\% for 3 times in a row. \newline ~\textbf{Rouge scores with Reference Summary (R-1/R-2/R-L)}: 0.78/ 0.63/ 0.78 \newline ~\textbf{Ours (\textit{RDASS})}: 0.44}              \\ \hline
    
    \multicolumn{2}{|l|}{\textbf{Correct Candidate}}            \\ \hline
    \multicolumn{2}{|L|}{ `슬기로운 의사생활' \textcolor{blue}{최고} 시청률 경신... 3주 연속 \textcolor{blue}{상승}세 \newline The tv program ``Sage doctor life'' set an all time record for its highest viewer ratings for 3 consecutive episodes. \newline ~\textbf{Rouge  scores with Reference Summary (R-1/R-2/R-L)}: 0.71/ 0.53/ 0.71 \newline ~\textbf{Ours (\textit{RDASS})}: 0.56}              \\ \hline
    
    \end{tabular}
    \end{adjustbox}
    \caption{An example showing the limitations of ROUGE in Korean summarization. The incorrectly generated summary has a high ROUGE score, but has the opposite semantic meaning. Text areas marked in blue and red serve as indicators for distinguishing the factualness of the semantic comparisons, as reflected by the our metrics shown.}
    \label{limit_table}
\end{table}

To overcome this limitation, an evaluation method that considers the semantic information of both the generated and reference summary is required. It is important to examine the factuality between the generated summary and source document, because the generated summary may contain false information. Each person summarizes information in different manners, and it is difficult to agree, even after cross-checking~\cite{kryscinski2019neural}. Therefore, the source document should also be considered with generated and reference summary.

In this study, we propose metrics for evaluating a summarization model that consider both the source document and reference summary together with the generated summary (see Table~\ref{limit_table}). Our contributions can be summarized as follows:
\begin{itemize}
\item We propose the evaluation metrics that can be applied to a summarization model using deep semantic information.
\item We propose methods to improve the correlation between the proposed evaluation metrics and human judgment.
\item Via extensive evaluation, we demonstrate that the correlation with human judgment is significantly higher for our proposed evaluation metrics than for ROUGE scores.

\end{itemize}

\section{Related Work}
Evaluation methods of text summarization are divided into two strategies: manual and automatic. Manual evaluation is expensive and difficult~\cite{nenkova2004evaluating,passonneau2013automated}. Several studies have been conducted to develop automatic methods that facilitate fast and low-cost evaluations. There are two types of automatic evaluation methods: extrinsic and intrinsic. An extrinsic automatic method evaluates a summarization model based on how it affects the completion of tasks comprising the judgment of document relevance \cite{dorr2004extrinsic}. The intrinsic automatic method evaluates quality via a property analysis or by calculating its similarity to a manually generated summary. Intrinsic methods include the pyramid method \cite{nenkova2007pyramid}, the basic-elements method \cite{hovy2006automated}, and ROUGE \cite{lin2004rouge}. The pyramid method inspects various human-made summaries and creates summary content units, each with a scoring weight. The basic-elements method is similar to the pyramid method. ROUGE evaluates the similarity of the lexical overlap between the candidate and reference summary. 

As the ROUGE score is computed based on the n-gram overlap, it does not account for synonymous words or phrases. Many approaches have been proposed to overcome this limitation. ParaEval \cite{zhou2006paraeval}, ROUGE-WE \cite{ng2015better}, ROUGE 2.0 \cite{ganesan2018rouge}, and ROUGE-G \cite{shafieibavani2018graph} have been used to extend ROUGE to support synonymous constructs. ParaEval uses a matching method based on paraphrase tables. ROUGE-WE uses a lexical matching method with a semantic similarity measure and the cosine distances between tokens. ROUGE 2.0 uses WordNet as a synonym dictionary and computed token overlaps with all synonyms of matched words. ROUGE-G uses lexical and semantic matching from WordNet. These approaches have limitations because they require hand-crafted lexical and synonym dictionaries, which are particularly difficult to construct in Korean. Our research is similar to \cite{zhang2019bertscore}, which utilized BERT to compute semantic score between the generated and reference sentence. However, \cite{zhang2019bertscore} does not consider the document, whereas our research considers the document to be characterized in the evaluation of summarization tasks. Overall, our research is different from previous approaches in that 1) We propose a method to evaluate generated summary by considering documents as well as reference summary. 2) In addition, our evaluation model is robust to out of vocabulary (OOV) words because it leverages a pre-trained neural network (SBERT) based on byte pair encoding (BPE)~\cite{gage1994new} tokenization method from unsupervised learning. Considering the fact that Korean is an agglutinative language, this feature is essential. 3) Finally, Our evaluation model can be further trained to capture more contextualized information both on reference summary and document.

Text summarization models can be divided into abstractive, extractive, and hybrid. Abstractive models reword phrases and create summaries having novel phrases constructed from the original document. Recent text summarization approaches have leveraged multi-task and multi-reward training \cite{jiang2018closed,paulus2017deep,pasunuru2018multi,guo2018soft}, attention-with-copying mechanisms \cite{tan2017abstractive,see2017get,cohan2018discourse}, and unsupervised training strategies \cite{schumann2018unsupervised,chu2018unsupervised}. The extractive method extracts the most-suitable sentences (or words) from the source document and copies them directly into the summary. Many researchers \cite{neto2002automatic,colmenares2015heads,filippova2013overcoming} have utilized domain expertise to develop heuristics for refining summary texts. Recently, neural-based text summarization models have been proposed to train the model for predicting whether a span of text should be included in the summary \cite{nallapati2016classify,narayan2017neural,xu2019neural,liu2019comes}. Reinforcement learning-based summarization models have also been proposed to directly optimize models \cite{wu2018learning,dong2018banditsum,narayan2018ranking}. The hybrid approach uses both abstractive and extractive methods. With this approach, the summarization process is divided into two phases: content selection and paraphrasing \cite{gehrmann2018bottom,hsu2018unified,chen2018fast,liu2018generating}.

\section{Methodology}
From Table~\ref{limit_table}, we can observe the importance of considering both the document and reference summary together for proper evaluation of the summarization model. In Subsection~\ref{metric_1}, we propose a method for evaluating the generated summary with the reference summary to reflect deep semantic meaning. Next, we propose a method for evaluating the generated summary with the original document and reference summary together. The reference-document-aware evaluation metric model can be further trained to capture more contextualized information from both on reference summary and document (Subsection~\ref{metric_learning}).

\subsection{Reference and Document Aware Semantic Evaluation}
\label{metric_1}
Let us define the generated summary from the summarization model as \(y_p=[w_1, ..., w_n]\) and reference summary as \(y_r=[w_1, ..., w_m] \), where \(w_i\) indicates each word. Then, each summary representation, \(v_p\) and \(v_r\), can be constructed using sentence-embedding methods. Neural-based sentence-embedding methods have been broadly studied. Conneau~\shortcite{conneau2017supervised} trained a siamese bidirectional long short-term memory model with a max-pooling strategy on the Stanford Natural Language Inference (SNLI) corpus~\cite{bowman2015large} and the MultiGenre Natural Language Inference (NLI) dataset~\cite{williams2017broad}. Cer~\shortcite{cer2018universal} proposed the universal sentence encoder to train a transformer on the SNLI dataset. Reimers~\shortcite{reimers2019sentence} recently proposed sentence-BERT(SBERT), which leverages a pre-trained BERT~\cite{devlin2018bert}, trained with a combination of the SNLI and multi-genre NLI, and showed state-of-the-art sentence embedding performance. SBERT is suitable for semantic similarity searches and showed faster inference speeds than previous state-of-the-art approaches, including BERT, RoBERTa~\cite{liu2019roberta}, and the universal sentence encoder. 

We leverage a pre-trained SBERT to construct summary representations. Each word representation, \(e\), is obtained from SBERT as
\begin{equation}
        E=[e_{cls},e_1, ..., e_n, e_{sep}]=\text{SBERT}([CLS], w_1, ..., w_n,.[SEP]) \label{eq:1} .\\
\end{equation}
Subsequently, mean-pooling is performed to construct \(v_p\) as,
\begin{equation}
    v_p(j) = \frac{\sum\limits_{i=0}^{n} e_i[j]}{n} \label{eq:2}
\end{equation}
where \(j\) represents an index of a word-embedding dimension, and \(n\) represents a length of \(E\). \(v_r\) can also be obtained in the same manner.

The semantic similarity score, \(s(p,r)\), between \(v_p\) and \(v_r\) can be obtained as follows:
\begin{equation}
s(p,r) = cos(v_p, v_r) = \frac{v_{p}^{\mathrm{T}}\cdot v_r}{\left \| v_p \right \|\left \| v_r \right \|} .
\end{equation}

Recall that it is important to consider factual consistency of generated summary with the source document, and, given the same document, the method of summarizing important information varies from person to person~\cite{owczarzak2012assessing,kryscinski2019neural}. Therefore, the source document should also be considered with the generated summary when evaluating the summarization model.

Given a document, \(D=[w_1, ..., w_k]\), the document representation, \(v_{d}\), can be obtained using Eqs. (1) and (2). Thus, the similarity score between \(v_{p}\) and \(v_d\) can be defined as,
\begin{equation}
s(p,d) = cos(v_{p}, v_d) = \frac{v_{p}^{\mathrm{T}}\cdot v_d}{\left \| v_{p} \right \|\left \| v_d \right \|} .
\end{equation}

Given a reference and source document, the reference-document-aware semantic score (\textit{RDASS}) of the generated summary is defined by averaging \(s(p,r)\) and \(s(p,d)\):
\begin{equation}
RDASS =  \frac{s(p,r) + s(p, d)}{2}.
\end{equation}
We also experimented with a sum, max and min operation between \(s(p,r)\) and \(s(p,d)\), but averaging the two scores reports highest correlation with human judgment.

\subsection{Fine-tuning SBERT with the Abstractive Summarization Model}
\label{metric_learning}
As SBERT is a trainable metric model, it can be further trained to capture more contextualized information about the reference summary and source document. We propose a fine-tuning method for SBERT that uses the abstractive summarization model. 

Most neural approaches for abstractive summarization are based on an encoder--decoder architecture~\cite{see2017get}. Formally, given a document, \(D=[w_1, ..., w_k]\), the objective is to generate a summary, \(y_p=[w_1, ..., w_n]\), from a hidden representation, \(h_p=[h_1, ..., h_n]\). The hidden representation is the output vector of the decoder. We leverage the hidden representation of the decoder to fine-tune the SBERT. 

Following~\cite{reimers2019sentence}, we adopt a triplet objective to fine-tune the SBERT. Given an anchor \(h_{p}\), a positive reference representation \(v_{r}^{p}\), a negative representation \(v_{r}^{n}\), and a Euclidean distance \(d\), the triplet objective for generated and reference summaries \(J(p,r)\) is then defined as
\begin{equation}
J(p,r) = max(0, \epsilon + d(h_p, v_{r}^{p}) - d(h_p, v_{r}^{n})),
\end{equation}
where \(\epsilon\) represents a margin that ensures \(h_p\) is closer to \(v_{r}^{p}\) than \(v_{r}^{n}\). We set \(\epsilon\) as \(1\). Similarly, the triplet objective for generated summary and document can be defined as
\begin{equation}
J(p,d) = max(0, \epsilon + d(h_p, v_{d}^{p}) - d(h_p, v_{d}^{n})).
\end{equation}
Thus, the final objective for SBERT is to minimize the combined two triplet objectives as
\begin{equation}
J = J(p,r) + J(p,d).
\end{equation}
The objective function SBERT \(J\) is jointly optimized with the abstractive summarization objective. Usually, the negative log-likelihood objective between the generated and reference summaries is used for abstractive summarization~\cite{see2017get,narayan2018don}. We refer to the fine-tuned SBERT with abstractive summarization model as ``FWA-SBERT.''

\section{Experimental Setup}
\subsection{Dataset}
We trained and evaluated our models using the Korean Daum/News dataset\footnote{https://media.daum.net/}, comprising 10 topics, such as politics, economy, international, culture, information technology, and others. From this, we extracted 3-million news articles. The number of articles for training, validating, and testing was \(2.98M\), \(0.01M\), and \(0.01M\) respectively. We refer to this dataset as Daum/News. We used Daum/News to fully understand the content of the article and conduct a proper evaluation. The dataset contains articles from 143 newspapers, each having different summary styles, and the effectiveness of the proposed methods is exemplified using it. Therefore, we expect that our research can be applied to different languages.

\subsection{Summarization Model}
We adopted abstractive summarization model of \cite{liu2019text}~\footnote{https://github.com/nlpyang/PreSumm}. Liu~\shortcite{liu2019text} leveraged pre-trained BERT as an encoder and a six-layered transformer as a decoder, showing state-of-the-art results on Cable News Network/DailyMail~\cite{hermann2015teaching}, New York Times~\cite{sandhaus2008new}, and XSum~\cite{narayan2018don} datasets. We set all environments according to~\cite{liu2019text}, except that we leveraged the pre-trained BERT trained on Korean dataset (Subsection \ref{sec:sbert}) instead of english-bert-base-uncased. We trained the abstractive summarization model on Korean Daum/News dataset.

\subsection{SBERT}
\label{sec:sbert}
To leverage SBERT, we first pre-trained BERT (bert-base-uncased) on Korean dataset, comprising \(23M\) sentences and \(1.6M\) documents, including Wiki, Sejong corpus, and web documents. Next, we trained SBERT with classification and regression objectives from NLI~\cite{bowman2015large,williams2017broad} and the semantical textual similarity (STS) benchmark (STSb)~\cite{cer2017semeval}. Because NLI and STSb datasets are in English, we leveraged the Korean NLI and STS dataset~\footnote{https://github.com/kakaobrain/KorNLUDatasets}~\cite{ham2020kornli} which translated from Kakao Machine Translator~\footnote{https://translate.kakao.com}. Evaluation of the STS benchmark test dataset was conducted, showing an \(80.52\) Spearman's rank correlation result. Subsequently, the pre-trained SBERT model was fine-tuned with the abstractive summarization model to capture more contextualized information of the reference summary and source document with a generated summary (Subsection \ref{metric_learning}). All training was conducted on the Kakao Brain Cloud with 4 Tesla V100 graphical processing units.

\subsection{Human Judgment}
To demonstrate the effectiveness of the reference-document-aware semantic metric, we evaluated its correlation with human judgment. Following \cite{kryscinski2019neural}, we asked annotators to score relevance, consistency, and fluency. \textbf{Relevance} represents the degree of appropriateness of the document, \textbf{consistency} represents the degree of factualness, and \textbf{fluency} represents the degree of the quality of generated summary. Additionally, \textbf{human avg} represents the average value of the scores for the three indicators. Given a document, reference summary, and generated summary, each annotator scored in the range of \(1\) to \(5\) points for the evaluation indicator (i.e., relevance, consistency, fluency). The human judgment was conducted by \(6\) judges having a PhD (3 judges) or a MS (3 judges) degree in computer science. The averaged human score of relevance was \(3.8\), consistency was \(3.6\), and fluency was \(3.9\) for \(200\) sampled summaries from Korean Daum/News test dataset.

\section{Results}
\label{sec:results}
In this section, we first report the performance of the summarization model using the ROUGE and proposed evaluation metrics (Subsection~\ref{metric_1}). Next, we report how the proposed evaluation metrics correlated to human judgment. We also report the correlation of the proposed evaluation metrics to ROUGE to show that the proposed methods complement ROUGE. Finally, through qualitative evaluation, we demonstrate the limitations of ROUGE and the superiority of the proposed evaluation metrics.

\subsection{Performance of the Summarization Model}

\begin{table}[h] \centering
\begin{adjustbox}{width=0.8\textwidth}
\begin{tabular}{ccccccc}
\multirow{2}{*}{Model} & \multicolumn{3}{c}{Proposed Evaluation Metrics} & \multicolumn{3}{c}{ROUGE} \\ 
                       & \(s(p,r)\)      & \(s(p,d)\)      & \textit{RDASS}      & 1       & 2      & L      \\ \hline
Reference Summary & 1.00 & 0.55 & 0.78 & 1.00 & 1.00 & 1.00 \\ \hline
Lead-1                 & 0.71        & 0.64        & 0.68                & 0.13    & 0.03   & 0.13   \\
Lead-3                 & 0.66        & 0.79        & 0.73                & 0.07    & 0.01   & 0.07   \\ \hline
BERTSUMABS \cite{liu2019text}                & 0.83        & 0.46        & 0.65                & 0.35    & 0.15   & 0.35
\end{tabular}
\end{adjustbox}
\caption{Performance of the summarization model on the DAUM/NEWS dataset.}
\label{performance}
\end{table}

The abstractive summarization model is based on the neural architecture of~\cite{liu2019text}. We trained the summarization model on the Daum/News dataset. To evaluate the summarization model, we used ROUGE and the proposed evaluation metrics. The fine-tuned FWA-SBERT was then used to evaluate the proposed semantic scores (\(s(p,r)\), \(s(p,d)\), and \textit{RDASS}). Table~\ref{performance} shows the performance of the summarization model with baseline methods (Reference Summary, Lead 1, and 3) on the Daum/News dataset. 

We set the reference summary as upper-bound. In the case of the reference summary, the reporter tends to use implicit words when summarizing the document, so the \(s(p,d)\) score is relatively low compared to the Lead baselines. However, because the \(s(p,r)\) score is 1.00, the reference summary shows the highest \textit{RDASS} score. For Lead-1, \(s(p,r)\) shows higher performance than \(s(p,d)\), and for Lead-3, \(s(p,d)\) shows higher performance than \(s(p,r)\). The reason for this performance is that Lead-3 contains more sentences from the document, so the similarity with the reference summary \(s(p,r)\) is low, but the similarity with the document \(s(p,d)\) is increased. In the case of ROUGE performance of lead baselines, relatively low performance can be confirmed compared to other researches~\cite{kryscinski2019neural} conducted in English dataset. The reason is that in the case of Korean, the same semantic meaning is expressed differently because of the nature of the language of the agglutinative language. A detailed example of this is described in Table~\ref{tab:example1} below. However, it can be seen that the \textit{RDASS} score of lead baselines is similar to that of the reference summary. Through this, we can confirm that the proposed evaluation method can reflect the semantic meaning of the reference summary and document well. In the case of the \cite{liu2019text}, it shows higher similarity with the reference summary than the Lead baselines, but since it is based on the generation model, it does not extract the sentence from the document as the Lead baselines. As a result, it shows the relatively low \(s(p,d)\) score. We describe how these results are correlated with human judgment in the next section.

\subsection{Correlation with Human Judgment}
\begin{figure*}[ht]
\centering
\begin{subfigure}{0.9\textwidth}\centering
\includegraphics[width=\columnwidth]{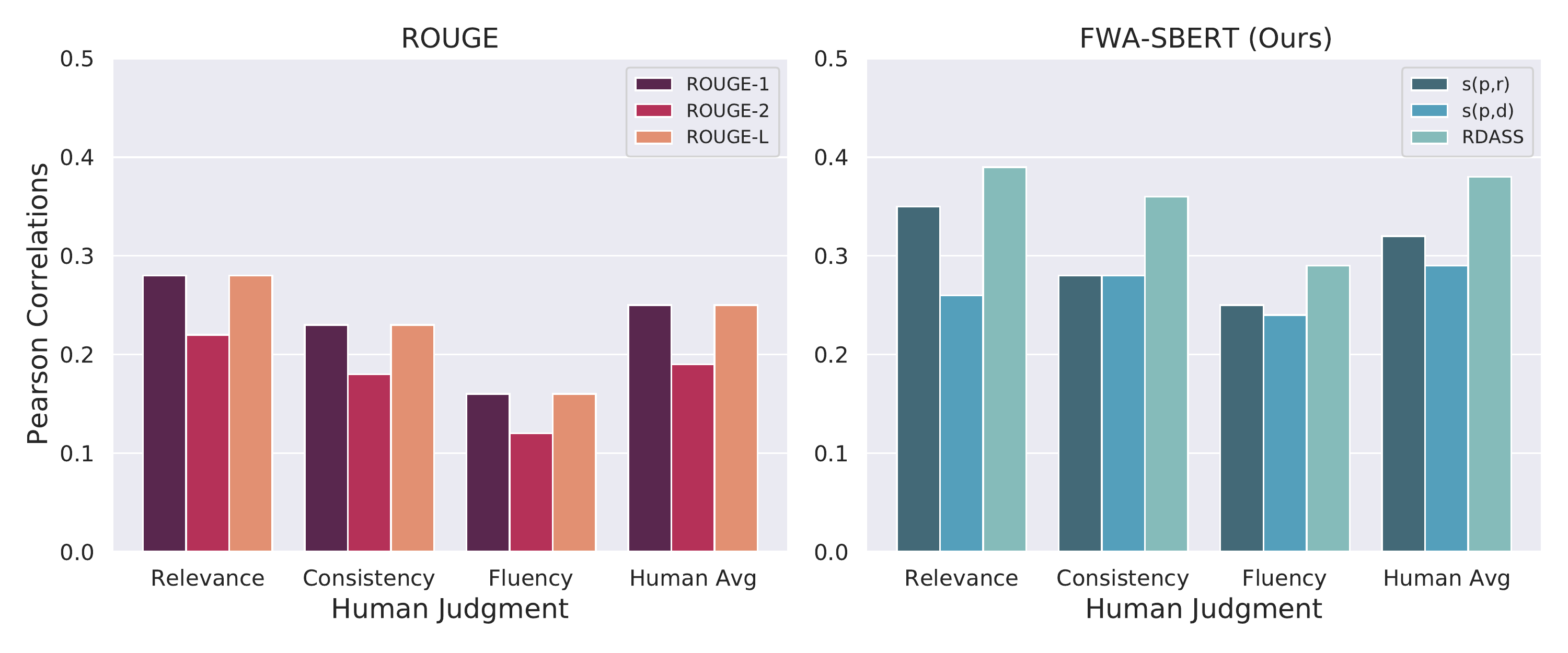}  
\caption{Pearson correlations} \label{fig:pearson_correlations} 
\end{subfigure}
\hfill
\begin{subfigure}{0.9\textwidth}\centering
\includegraphics[width=\columnwidth]{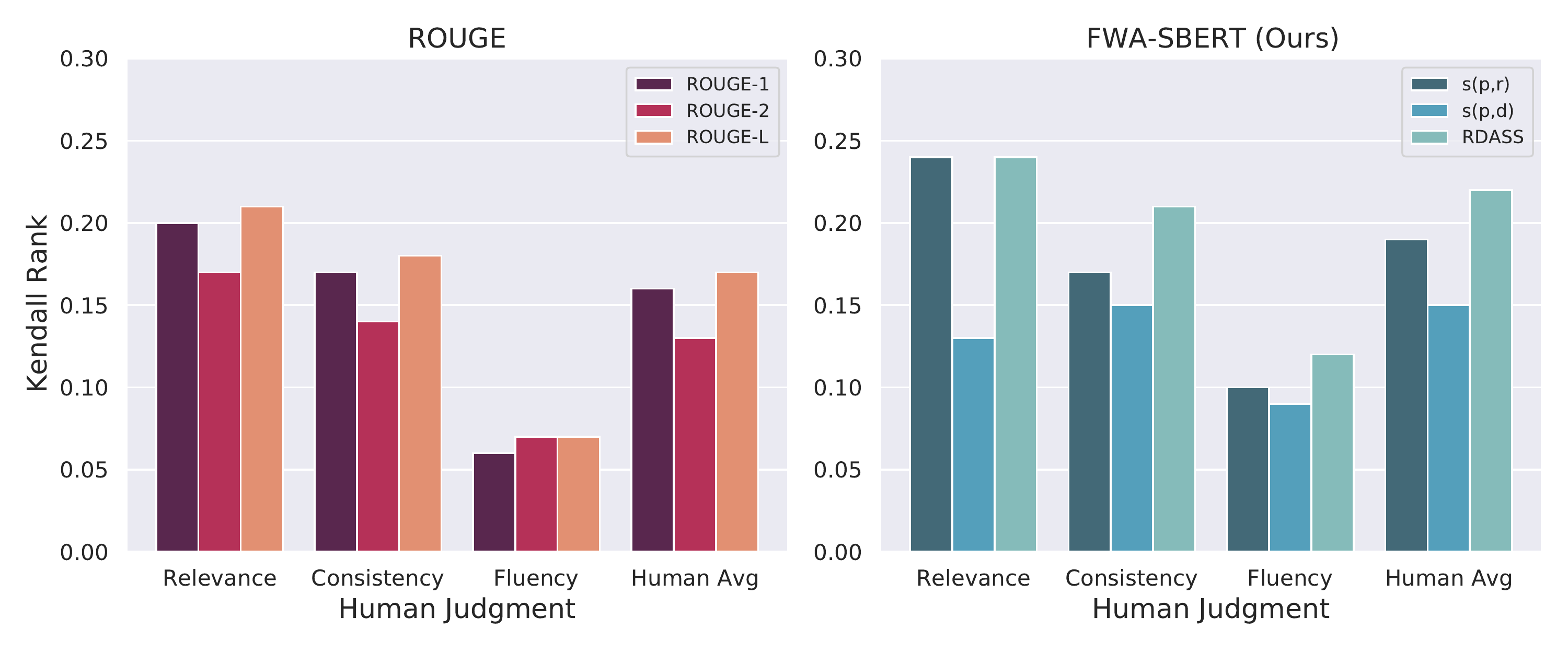}  
\caption{Kendall rank} \label{fig:kendall_rank}
\end{subfigure}
\label{fig}
\caption{Pearson correlations and Kendall rank of the proposed evaluation metrics with human judgment.}
\end{figure*}

Figures (\ref{fig:pearson_correlations}) and (\ref{fig:kendall_rank}) show the Pearson correlation and Kendall rank, respectively, of the proposed evaluation metrics with human judgment on the 200 sampled summaries. Pearson correlation measure whether the two variables are linearly related, where 1 indicates positive linear correlation and -1 indicates negative linear correlation. And Kendall rank measure the rank correlation of the two variables, where 1 indicates two variables are similar and -1 indicates dissimilar. Both correlation measure methods are widely used in summarization task to analyze correlation with human judgment. 

In the Pearson correlation matrix, the correlation with human judgment was significantly higher for the proposed evaluation metrics than for ROUGE scores. Additionally, in the Kendall rank matrix, the proposed evaluation metrics showed highest correlation with human judgment than did the ROUGE scores. Among the proposed evaluation metrics, \(s(p, r)\) showed higher performance than \(s(p, d)\) and \textit{RDASS} showed the highest correlation with human judgment. These results indicate that the proposed evaluation metrics can reflect deep semantic meaning overcoming the limitations of ROUGE which based on n-gram overlap.

\input{tables/table3_Pearson_Correlations}

To demonstrate the effectiveness of fine-tuning SBERT with an abstractive summarization model, we set baseline methods depending on which sentence representation methods to use for the proposed methods (Subsection ~\ref{metric_1}) as follows:

\textbf{Multilingual Universal Sentence Encoder (MUSE)}: MUSE~\cite{yang2019multilingual} is a multilingual sentence encoder that embeds text from 16 languages into a single semantic space using multi-task learning. This model was trained on more than 1-billion question-answer pairs and showed competitive state-of-the-art results on semantic~\cite{gillick2018end}, bitext retrival~\cite{ziemski2016united}, and retrieval question-answering~\cite{yang2019multilingual}. 

\textbf{Pre-trained SBERT}: We only leveraged pre-trained SBERT without fine-tuning. We refer to this as ``P-SBERT.''

Table \ref{tab:pearson} show the performance comparison depended upon which sentence representation was used. P-SBERT shows the high correlation coefficient with humans than MUSE. Overall, when the FWA-SBERT was used, it showed the closest correlation with human judgment. 

Through quantitative evaluation, we demonstrated that the proposed evaluation metrics had a high correlation with human judgment and that the method of fine-tuning SBERT improved the performance of the proposed evaluation metrics.

We also experimented to understand how each evaluation metric was correlated to each other. As shown in Table~\ref{tab:pearson2}, there was a high correlation among the ROUGE metrics. However, the proposed evaluation metrics had a relatively low correlation with ROUGE. This indicates that the proposed evaluation metrics reflected semantic meaning, in our case, that ROUGE could not. Thus, we believe it complements ROUGE metrics.

\begin{table}[h] \centering
\begin{tabular}{c|cccccc}
         & ROUGE-1 & ROUGE-2 & ROUGE-L & \(s(p,r)\) & \(s(p,d)\) & \textit{RDASS} \\ \hline
ROUGE-1  & 1.00    & 0.84    & 0.99    & 0.64   & 0.16   & 0.54     \\
ROUGE-2  &         & 1.00    & 0.85    & 0.52   & 0.09   & 0.45     \\
ROUGE-L  &         &         & 1.00    & 0.63   & 0.17   & 0.53     \\
\(s(p,r)\)   &         &         &         & 1.00   & 0.32   & 0.77     \\
\(s(p,d)\)   &         &         &         &        & 1.00   & 0.69     \\
\textit{RDASS} &         &         &         &        &        & 1.00    
\end{tabular}
\caption{Pearson correlation of ROUGE and the proposed evaluation metrics.}
\label{tab:pearson2}
\end{table}

\subsection{Qualitative Analysis}
In this section, through qualitative analysis, we demonstrate the effectiveness of our evaluation metrics. Table~\ref{tab:example1} shows ROUGE, RDASS and human evaluation results for the generated summaries for the two articles. 

\begin{table}[t]
\begin{adjustbox}{width=1.0\textwidth}
    \centering
    \begin{tabular}{|l|l|}
    \hline
    \multicolumn{2}{|l|}{\textbf{Article-1}}            \\ \hline
    \multicolumn{2}{|L|}{ 리오넬 메시(30·fc바르셀로나)가 자신의 서른 번째 생일을 가족과 함께 오붓하게 보냈다. 지난 24일 만 서른 살이 된 메시는 자신의 인스타그램에 집에서 가족들과 함께 보낸 생일상을 찍은 사진을 올렸다. 메시는 오랜 그의 여자친구이자, 이제 아내가 되는 안토넬라 로쿠조(29), 아들 티아고가 함께 다정하게 사진을 찍었다. \newline Lionel Messi (30 fc Barcelona) spent his thirtieth birthday with his family. Messi, who turned thirty on the 24th, posted a picture of his birthday on Instagram with his family at home. Messi was tenderly photographed by his longtime girlfriend, Antonella Rokujo (29), and his son, Thiago.}              \\ \hline
    \multicolumn{2}{|l|}{\textbf{Reference Summary}}           \\ \hline
    \multicolumn{2}{|L|}{ 메시가 30번째 생일 함께한 이는 아내와 아들 \newline Messi's 30th birthday with his wife and son.}              \\ \hline
    
    \multicolumn{2}{|l|}{\textbf{Generated Summary}}            \\ \hline
    \multicolumn{2}{|L|}{ 메시 30번째 생일, 가족과 함께 오붓하게 보내 \newline On the 30th birthday of Messi, he had a good time with his family. \newline ~\textbf{Rouge(1/ 2/ L)}: 0.14/ 0.00/ 0.14 \newline ~\textbf{\textit{RDASS}}: 0.81 \newline ~\textbf{Human Evaluation (relevance/ consistency/ fluency)}: 4.4/ 4.2/ 4.2}              \\ \hline
    
    \multicolumn{2}{|l|}{\textbf{Article-2}}            \\ \hline
    \multicolumn{2}{|L|}{삼성전자는 19일(현지시간) 브라질 상파울루의 팔라시오 탕가라 호텔에서 `QLED TV 론칭 이벤트'를 열고 2017년형 QLED TV 라인업을 선보였다고 23일 밝혔다. 4월 중남미에서는 처음으로 멕시코에서 QLED TV를 출시한 뒤 파나마, 콜롬비아 등으로 확대하다 이번에 중남미 최대 시장인 브라질에 제품을 출시한 것이다. 브라질은 전체 중남미 TV 시장의 40\%(금액 기준) 이상을 차지할 정도로 중요한 TV 시장이다. 올해 1∼4월 브라질 TV 시장은 작년 같은 기간보다 13\%(수량 기준) 성장했고, 특히 프리미엄 TV 시장인 UHD(초고화질) TV는 작년보다 50\% 이상 시장이 커졌다. 특히 삼성전자는 브라질 UHD TV 시장에서 올해 1∼4월 56\%(수량 기준) 점유율로 압도적 1위를 차지했다. \newline Samsung Electronics announced on the 23rd that it held a `QLED TV launching event' at the Palacio Tangara Hotel in Sao Paulo, Brazil on the 19th (local time) and introduced the 2017 QLED TV lineup. In April, it launched the QLED TV in Mexico for the first time in Latin America, and then expanded to Panama, Colombia, etc. This time it launched the product in Brazil, the largest market in Latin America. Brazil is an important TV market, accounting for more than 40\% of total Latin American TV market. In January-April this year, the Brazilian TV market grew 13\% (in quantity) from the same period last year. In particular, the UHD (Ultra High Definition) TV, a premium TV market, was 50\% larger than last year. In particular, Samsung Electronics took the dominant position in the UHD TV market in Brazil with 56\% (based on quantity) in January-April this year.}              \\ \hline
    \multicolumn{2}{|l|}{\textbf{Reference Summary}}            \\ \hline
    \multicolumn{2}{|L|}{ 삼성전자, 중남미 최대 시장 브라질에 qled tv 론칭 \newline Samsung Electronics launches `qled tv' in Brazil, the largest market in Latin America.}              \\ \hline
    
    \multicolumn{2}{|l|}{\textbf{Generated Summary}}            \\ \hline
    \multicolumn{2}{|L|}{ 삼성전자, 브라질서 `qled tv' 신제품 출시 \newline Samsung Electronics launches new `qled tv' in Brazil. \newline ~\textbf{Rouge(1/ 2/ L)}: 0.14/ 0.00/ 0.14 \newline ~\textbf{\textit{RDASS}}: 0.71 \newline ~\textbf{Human Evaluation(relevance/ consistency/ fluency)}: 4.6/ 4.4/ 4.4}              \\ \hline

    \end{tabular}
    \end{adjustbox}
    \caption{Example articles from the DAUM/News test dataset. ROUGE, RDASS and human evaluation results for the generated summaries are represented.}
    \label{tab:example1}
\end{table}

In article-1, the generated summary \textit{``On the 30th birthday of Messi, he had a good time with his family"} has the same semantic meaning as the reference summary \textit{``Messi's 30th birthday with his wife and son"}. However, since the sentence having the same semantic meaning can be variously expressed in Korean, which has the characteristics of agglutinative language, the ROUGE score is low while human evaluation scores are high. Likewise, the generated summary \textit{``Samsung Electronics launches new ‘qled tv’ in Brazil”} in article-2 has a same semantic meaning as the reference summary \textit{``Samsung Electronics launches ‘qled tv’ in Brazil, the largest market in Latin America”}. The generated summary in both articles is correct, but the ROUGE score is low. On the other hand, the \textit{RDASS} score indicates a higher score, and indicates that the generated summary is the correct answer.

\section{Conclusion}

In this paper, we pointed out the limitation of the widely used ROUGE evaluation metric when adopting Korean summarization. Since Korean is an agglutinative language, the generated summary having the same semantic meaning with reference summary can be variously expressed. Therefore, only leveraging ROUGE metric can produce inaccurate evaluation results. To overcome this limitation, we proposed \textit{RDASS (\textbf{R}eference and ~\textbf{D}ocument ~\textbf{A}ware ~\textbf{S}emantic ~\textbf{S}core)} evaluation metric. The \textit{RDASS} can reflect deep semantic relationships of a generated, reference summary, and document. Through extensive evaluations, we demonstrated that the correlation with human judgment is higher for the proposed evaluation metric (\textit{RDASS}) than for ROUGE scores. In future work, we will demonstrate the effectiveness of the proposed method in English summarization dataset.

\section*{Acknowledgements}
This research was supported by the MSIT(Ministry of Science and ICT), Korea, under the ITRC(Information Technology Research Center) support program (IITP-2020-2018-0-01405) supervised by the IITP(Institute for Information \& Communications Technology Planning \& Evaluation).
Also, thanks to 'GyoungEun Han' of the Kakao Enterprise NLP team for giving linguistic advice in writing this paper.

\bibliographystyle{coling}
\bibliography{coling2020}

\end{document}

%% file: tables/table3_Pearson_Correlations.tex
\begin{table*}[h]\centering
\begin{adjustbox}{width=0.9\textwidth}

\begin{tabular}{cc|cccccccc}

\toprule
\multicolumn{2}{c|}{\multirow{2}{*}{Sentence Representation}}
& \multicolumn{2}{c|}{Relevance} & \multicolumn{2}{c|}{Consistency} 
& \multicolumn{2}{c|}{Fluency} & \multicolumn{2}{c}{Human Avg} \\ 
\cmidrule{3-10} 
 \multicolumn{2}{c|}{} & Pearson & \multicolumn{1}{c|}{Kendall}  & Pearson & \multicolumn{1}{c|}{Kendall} & Pearson & \multicolumn{1}{c|}{Kendall} & Pearson & \multicolumn{1}{c}{Kendall}\\
\midrule
\multirow{3}{*}{MUSE} &\(s(p,r)\) &0.29 & \multicolumn{1}{c|}{0.19} & 0.18 & 
\multicolumn{1}{c|}{0.10} & 0.22& \multicolumn{1}{c|}{0.08} & 0.25 &0.13\\
&\(s(p,d)\) &0.09 & \multicolumn{1}{c|}{0.05} & 0.13 & \multicolumn{1}{c|}{0.06}& 0.15& \multicolumn{1}{c|}{0.04}& 0.13 & 0.06\\
&\textit{RDASS} &0.29 &\multicolumn{1}{c|}{0.19} & 0.24 & \multicolumn{1}{c|}{0.12}& 0.23 &\multicolumn{1}{c|}{0.09} & 0.28 &0.14 \\
\midrule
\multirow{3}{*}{P-SBERT} &\(s(p,r)\) &0.34 & \multicolumn{1}{c|}{0.22}& 0.27 &\multicolumn{1}{c|}{0.17} & 0.25 & \multicolumn{1}{c|}{0.09}& 0.32 &0.18\\
&\(s(p,d)\) &0.24 & \multicolumn{1}{c|}{0.13}& 0.27 & \multicolumn{1}{c|}{0.15}& 0.22 & \multicolumn{1}{c|}{0.09}& 0.27 & 0.15\\
&\textit{RDASS} &0.37 & \multicolumn{1}{c|}{0.22}& 0.34 & \multicolumn{1}{c|}{0.20}& 0.29 & \multicolumn{1}{c|}{0.11}& 0.37 &0.21\\
\midrule
\multirow{3}{*}{FWA-SBERT} &\(s(p,r)\) &0.35 & \multicolumn{1}{c|}{0.24}& 0.28 & \multicolumn{1}{c|}{0.17}& 0.25 & \multicolumn{1}{c|}{0.10}& 0.32 &0.19\\
&\(s(p,d)\) &0.26 & \multicolumn{1}{c|}{0.13}& 0.28 & \multicolumn{1}{c|}{0.15}& 0.24& \multicolumn{1}{c|}{0.09}& 0.29 & 0.15\\
&\textit{RDASS} &\textbf{0.39} & \multicolumn{1}{c|}{\textbf{0.24}}& \textbf{0.36} & \multicolumn{1}{c|}{\textbf{0.21}}& \textbf{0.29} & \multicolumn{1}{c|}{\textbf{0.12}} & \textbf{0.38} & \textbf{0.22}\\
\midrule

\end{tabular}  
\end{adjustbox}
\caption{Performance comparison depended upon which sentence representation was used.}
\label{tab:pearson}
\end{table*}

%% file: coling2020.bbl
\begin{thebibliography}{}

\bibitem[\protect\citename{Bowman \bgroup et al.\egroup }2015]{bowman2015large}
Samuel~R Bowman, Gabor Angeli, Christopher Potts, and Christopher~D Manning.
\newblock 2015.
\newblock A large annotated corpus for learning natural language inference.
\newblock {\em arXiv preprint arXiv:1508.05326}.

\bibitem[\protect\citename{Cer \bgroup et al.\egroup }2017]{cer2017semeval}
Daniel Cer, Mona Diab, Eneko Agirre, Inigo Lopez-Gazpio, and Lucia Specia.
\newblock 2017.
\newblock Semeval-2017 task 1: Semantic textual similarity-multilingual and
  cross-lingual focused evaluation.
\newblock {\em arXiv preprint arXiv:1708.00055}.

\bibitem[\protect\citename{Cer \bgroup et al.\egroup }2018]{cer2018universal}
Daniel Cer, Yinfei Yang, Sheng-yi Kong, Nan Hua, Nicole Limtiaco, Rhomni~St
  John, Noah Constant, Mario Guajardo-Cespedes, Steve Yuan, Chris Tar, et~al.
\newblock 2018.
\newblock Universal sentence encoder.
\newblock {\em arXiv preprint arXiv:1803.11175}.

\bibitem[\protect\citename{Chen and Bansal}2018]{chen2018fast}
Yen-Chun Chen and Mohit Bansal.
\newblock 2018.
\newblock Fast abstractive summarization with reinforce-selected sentence
  rewriting.
\newblock {\em arXiv preprint arXiv:1805.11080}.

\bibitem[\protect\citename{Chu and Liu}2018]{chu2018unsupervised}
Eric Chu and Peter~J Liu.
\newblock 2018.
\newblock Unsupervised neural multi-document abstractive summarization.
\newblock {\em arXiv preprint arXiv:1810.05739}.

\bibitem[\protect\citename{Cohan \bgroup et al.\egroup
  }2018]{cohan2018discourse}
Arman Cohan, Franck Dernoncourt, Doo~Soon Kim, Trung Bui, Seokhwan Kim, Walter
  Chang, and Nazli Goharian.
\newblock 2018.
\newblock A discourse-aware attention model for abstractive summarization of
  long documents.
\newblock {\em arXiv preprint arXiv:1804.05685}.

\bibitem[\protect\citename{Colmenares \bgroup et al.\egroup
  }2015]{colmenares2015heads}
Carlos~A Colmenares, Marina Litvak, Amin Mantrach, and Fabrizio Silvestri.
\newblock 2015.
\newblock Heads: Headline generation as sequence prediction using an abstract
  feature-rich space.

\bibitem[\protect\citename{Conneau \bgroup et al.\egroup
  }2017]{conneau2017supervised}
Alexis Conneau, Douwe Kiela, Holger Schwenk, Loic Barrault, and Antoine Bordes.
\newblock 2017.
\newblock Supervised learning of universal sentence representations from
  natural language inference data.
\newblock {\em arXiv preprint arXiv:1705.02364}.

\bibitem[\protect\citename{Devlin \bgroup et al.\egroup }2018]{devlin2018bert}
Jacob Devlin, Ming-Wei Chang, Kenton Lee, and Kristina Toutanova.
\newblock 2018.
\newblock Bert: Pre-training of deep bidirectional transformers for language
  understanding.
\newblock {\em arXiv preprint arXiv:1810.04805}.

\bibitem[\protect\citename{Dong \bgroup et al.\egroup }2018]{dong2018banditsum}
Yue Dong, Yikang Shen, Eric Crawford, Herke van Hoof, and Jackie Chi~Kit
  Cheung.
\newblock 2018.
\newblock Banditsum: Extractive summarization as a contextual bandit.
\newblock {\em arXiv preprint arXiv:1809.09672}.

\bibitem[\protect\citename{Dorr \bgroup et al.\egroup }2004]{dorr2004extrinsic}
Bonnie Dorr, Christof Monz, Douglas Oard, David Zajic, and Richard Schwartz.
\newblock 2004.
\newblock Extrinsic evaluation of automatic metrics for summarization.
\newblock Technical report, MARYLAND UNIV COLLEGE PARK INST FOR ADVANCED
  COMPUTER STUDIES.

\bibitem[\protect\citename{Filippova and Altun}2013]{filippova2013overcoming}
Katja Filippova and Yasemin Altun.
\newblock 2013.
\newblock Overcoming the lack of parallel data in sentence compression.
\newblock In {\em Proceedings of the 2013 Conference on Empirical Methods in
  Natural Language Processing}, pages 1481--1491.

\bibitem[\protect\citename{Gage}1994]{gage1994new}
Philip Gage.
\newblock 1994.
\newblock A new algorithm for data compression.
\newblock {\em C Users Journal}, 12(2):23--38.

\bibitem[\protect\citename{Ganesan}2018]{ganesan2018rouge}
Kavita Ganesan.
\newblock 2018.
\newblock Rouge 2.0: Updated and improved measures for evaluation of
  summarization tasks.
\newblock {\em arXiv preprint arXiv:1803.01937}.

\bibitem[\protect\citename{Gehrmann \bgroup et al.\egroup
  }2018]{gehrmann2018bottom}
Sebastian Gehrmann, Yuntian Deng, and Alexander~M Rush.
\newblock 2018.
\newblock Bottom-up abstractive summarization.
\newblock {\em arXiv preprint arXiv:1808.10792}.

\bibitem[\protect\citename{Gillick \bgroup et al.\egroup }2018]{gillick2018end}
Daniel Gillick, Alessandro Presta, and Gaurav~Singh Tomar.
\newblock 2018.
\newblock End-to-end retrieval in continuous space.
\newblock {\em arXiv preprint arXiv:1811.08008}.

\bibitem[\protect\citename{Guo \bgroup et al.\egroup }2018]{guo2018soft}
Han Guo, Ramakanth Pasunuru, and Mohit Bansal.
\newblock 2018.
\newblock Soft layer-specific multi-task summarization with entailment and
  question generation.
\newblock {\em arXiv preprint arXiv:1805.11004}.

\bibitem[\protect\citename{Ham \bgroup et al.\egroup }2020]{ham2020kornli}
Jiyeon Ham, Yo~Joong Choe, Kyubyong Park, Ilji Choi, and Hyungjoon Soh.
\newblock 2020.
\newblock Kornli and korsts: New benchmark datasets for korean natural language
  understanding.
\newblock {\em arXiv preprint arXiv:2004.03289}.

\bibitem[\protect\citename{Hermann \bgroup et al.\egroup
  }2015]{hermann2015teaching}
Karl~Moritz Hermann, Tomas Kocisky, Edward Grefenstette, Lasse Espeholt, Will
  Kay, Mustafa Suleyman, and Phil Blunsom.
\newblock 2015.
\newblock Teaching machines to read and comprehend.
\newblock In {\em Advances in neural information processing systems}, pages
  1693--1701.

\bibitem[\protect\citename{Hovy \bgroup et al.\egroup }2006]{hovy2006automated}
Eduard~H Hovy, Chin-Yew Lin, Liang Zhou, and Junichi Fukumoto.
\newblock 2006.
\newblock Automated summarization evaluation with basic elements.
\newblock In {\em LREC}, volume~6, pages 899--902. Citeseer.

\bibitem[\protect\citename{Hsu \bgroup et al.\egroup }2018]{hsu2018unified}
Wan-Ting Hsu, Chieh-Kai Lin, Ming-Ying Lee, Kerui Min, Jing Tang, and Min Sun.
\newblock 2018.
\newblock A unified model for extractive and abstractive summarization using
  inconsistency loss.
\newblock {\em arXiv preprint arXiv:1805.06266}.

\bibitem[\protect\citename{Jiang and Bansal}2018]{jiang2018closed}
Yichen Jiang and Mohit Bansal.
\newblock 2018.
\newblock Closed-book training to improve summarization encoder memory.
\newblock {\em arXiv preprint arXiv:1809.04585}.

\bibitem[\protect\citename{Kryscinski \bgroup et al.\egroup
  }2019]{kryscinski2019neural}
Wojciech Kryscinski, Nitish~Shirish Keskar, Bryan McCann, Caiming Xiong, and
  Richard Socher.
\newblock 2019.
\newblock Neural text summarization: A critical evaluation.
\newblock In {\em Proceedings of the 2019 Conference on Empirical Methods in
  Natural Language Processing and the 9th International Joint Conference on
  Natural Language Processing (EMNLP-IJCNLP)}, pages 540--551.

\bibitem[\protect\citename{Lin}2004a]{lin-2004-rouge}
Chin-Yew Lin.
\newblock 2004a.
\newblock {ROUGE}: A package for automatic evaluation of summaries.
\newblock In {\em Text Summarization Branches Out}, pages 74--81, Barcelona,
  Spain, July. Association for Computational Linguistics.

\bibitem[\protect\citename{Lin}2004b]{lin2004rouge}
Chin-Yew Lin.
\newblock 2004b.
\newblock Rouge: A packagefor automatic evaluation of summaries.
\newblock In {\em ProceedingsofWorkshop on Text Summarization Branches Out,
  Post2Conference Workshop of ACL}.

\bibitem[\protect\citename{Liu and Lapata}2019]{liu2019text}
Yang Liu and Mirella Lapata.
\newblock 2019.
\newblock Text summarization with pretrained encoders.
\newblock {\em arXiv preprint arXiv:1908.08345}.

\bibitem[\protect\citename{Liu \bgroup et al.\egroup }2018]{liu2018generating}
Peter~J Liu, Mohammad Saleh, Etienne Pot, Ben Goodrich, Ryan Sepassi, Lukasz
  Kaiser, and Noam Shazeer.
\newblock 2018.
\newblock Generating wikipedia by summarizing long sequences.
\newblock {\em arXiv preprint arXiv:1801.10198}.

\bibitem[\protect\citename{Liu \bgroup et al.\egroup }2019a]{liu2019comes}
Jingyun Liu, Jackie~CK Cheung, and Annie Louis.
\newblock 2019a.
\newblock What comes next? extractive summarization by next-sentence
  prediction.
\newblock {\em arXiv preprint arXiv:1901.03859}.

\bibitem[\protect\citename{Liu \bgroup et al.\egroup }2019b]{liu2019roberta}
Yinhan Liu, Myle Ott, Naman Goyal, Jingfei Du, Mandar Joshi, Danqi Chen, Omer
  Levy, Mike Lewis, Luke Zettlemoyer, and Veselin Stoyanov.
\newblock 2019b.
\newblock Roberta: A robustly optimized bert pretraining approach.
\newblock {\em arXiv preprint arXiv:1907.11692}.

\bibitem[\protect\citename{Louis and Nenkova}2013]{louis2013automatically}
Annie Louis and Ani Nenkova.
\newblock 2013.
\newblock Automatically assessing machine summary content without a gold
  standard.
\newblock {\em Computational Linguistics}, 39(2):267--300.

\bibitem[\protect\citename{Nallapati \bgroup et al.\egroup
  }2016]{nallapati2016classify}
Ramesh Nallapati, Bowen Zhou, and Mingbo Ma.
\newblock 2016.
\newblock Classify or select: Neural architectures for extractive document
  summarization.
\newblock {\em arXiv preprint arXiv:1611.04244}.

\bibitem[\protect\citename{Narayan \bgroup et al.\egroup
  }2017]{narayan2017neural}
Shashi Narayan, Nikos Papasarantopoulos, Shay~B Cohen, and Mirella Lapata.
\newblock 2017.
\newblock Neural extractive summarization with side information.
\newblock {\em arXiv preprint arXiv:1704.04530}.

\bibitem[\protect\citename{Narayan \bgroup et al.\egroup
  }2018a]{narayan2018don}
Shashi Narayan, Shay~B Cohen, and Mirella Lapata.
\newblock 2018a.
\newblock Don't give me the details, just the summary! topic-aware
  convolutional neural networks for extreme summarization.
\newblock {\em arXiv preprint arXiv:1808.08745}.

\bibitem[\protect\citename{Narayan \bgroup et al.\egroup
  }2018b]{narayan2018ranking}
Shashi Narayan, Shay~B Cohen, and Mirella Lapata.
\newblock 2018b.
\newblock Ranking sentences for extractive summarization with reinforcement
  learning.
\newblock {\em arXiv preprint arXiv:1802.08636}.

\bibitem[\protect\citename{Nenkova and Passonneau}2004]{nenkova2004evaluating}
Ani Nenkova and Rebecca~J Passonneau.
\newblock 2004.
\newblock Evaluating content selection in summarization: The pyramid method.
\newblock In {\em Proceedings of the human language technology conference of
  the north american chapter of the association for computational linguistics:
  Hlt-naacl 2004}, pages 145--152.

\bibitem[\protect\citename{Nenkova \bgroup et al.\egroup
  }2007]{nenkova2007pyramid}
Ani Nenkova, Rebecca Passonneau, and Kathleen McKeown.
\newblock 2007.
\newblock The pyramid method: Incorporating human content selection variation
  in summarization evaluation.
\newblock {\em ACM Transactions on Speech and Language Processing (TSLP)},
  4(2):4--es.

\bibitem[\protect\citename{Neto \bgroup et al.\egroup }2002]{neto2002automatic}
Joel~Larocca Neto, Alex~A Freitas, and Celso~AA Kaestner.
\newblock 2002.
\newblock Automatic text summarization using a machine learning approach.
\newblock In {\em Brazilian symposium on artificial intelligence}, pages
  205--215. Springer.

\bibitem[\protect\citename{Ng and Abrecht}2015]{ng2015better}
Jun-Ping Ng and Viktoria Abrecht.
\newblock 2015.
\newblock Better summarization evaluation with word embeddings for rouge.
\newblock {\em arXiv preprint arXiv:1508.06034}.

\bibitem[\protect\citename{Owczarzak and Dang}2011]{owczarzak2011overview}
Karolina Owczarzak and Hoa~Trang Dang.
\newblock 2011.
\newblock Overview of the tac 2011 summarization track: Guided task and aesop
  task.
\newblock In {\em Proceedings of the Text Analysis Conference (TAC 2011),
  Gaithersburg, Maryland, USA, November}.

\bibitem[\protect\citename{Owczarzak \bgroup et al.\egroup
  }2012]{owczarzak2012assessing}
Karolina Owczarzak, Hoa~Trang Dang, Peter~A Rankel, and John~M Conroy.
\newblock 2012.
\newblock Assessing the effect of inconsistent assessors on summarization
  evaluation.
\newblock In {\em Proceedings of the 50th Annual Meeting of the Association for
  Computational Linguistics: Short Papers-Volume 2}, pages 359--362.
  Association for Computational Linguistics.

\bibitem[\protect\citename{Passonneau \bgroup et al.\egroup
  }2013]{passonneau2013automated}
Rebecca~J Passonneau, Emily Chen, Weiwei Guo, and Dolores Perin.
\newblock 2013.
\newblock Automated pyramid scoring of summaries using distributional
  semantics.
\newblock In {\em Proceedings of the 51st Annual Meeting of the Association for
  Computational Linguistics (Volume 2: Short Papers)}, pages 143--147.

\bibitem[\protect\citename{Pasunuru and Bansal}2018]{pasunuru2018multi}
Ramakanth Pasunuru and Mohit Bansal.
\newblock 2018.
\newblock Multi-reward reinforced summarization with saliency and entailment.
\newblock {\em arXiv preprint arXiv:1804.06451}.

\bibitem[\protect\citename{Paulus \bgroup et al.\egroup }2017]{paulus2017deep}
Romain Paulus, Caiming Xiong, and Richard Socher.
\newblock 2017.
\newblock A deep reinforced model for abstractive summarization.
\newblock {\em arXiv preprint arXiv:1705.04304}.

\bibitem[\protect\citename{Reimers and Gurevych}2019]{reimers2019sentence}
Nils Reimers and Iryna Gurevych.
\newblock 2019.
\newblock Sentence-bert: Sentence embeddings using siamese bert-networks.
\newblock {\em arXiv preprint arXiv:1908.10084}.

\bibitem[\protect\citename{Sandhaus}2008]{sandhaus2008new}
Evan Sandhaus.
\newblock 2008.
\newblock The new york times annotated corpus.
\newblock {\em Linguistic Data Consortium, Philadelphia}, 6(12):e26752.

\bibitem[\protect\citename{Schumann}2018]{schumann2018unsupervised}
Raphael Schumann.
\newblock 2018.
\newblock Unsupervised abstractive sentence summarization using length
  controlled variational autoencoder.
\newblock {\em arXiv preprint arXiv:1809.05233}.

\bibitem[\protect\citename{See \bgroup et al.\egroup }2017]{see2017get}
Abigail See, Peter~J Liu, and Christopher~D Manning.
\newblock 2017.
\newblock Get to the point: Summarization with pointer-generator networks.
\newblock {\em arXiv preprint arXiv:1704.04368}.

\bibitem[\protect\citename{ShafieiBavani \bgroup et al.\egroup
  }2018]{shafieibavani2018graph}
Elaheh ShafieiBavani, Mohammad Ebrahimi, Raymond Wong, and Fang Chen.
\newblock 2018.
\newblock A graph-theoretic summary evaluation for rouge.
\newblock In {\em Proceedings of the 2018 Conference on Empirical Methods in
  Natural Language Processing}, pages 762--767.

\bibitem[\protect\citename{Tan \bgroup et al.\egroup }2017]{tan2017abstractive}
Jiwei Tan, Xiaojun Wan, and Jianguo Xiao.
\newblock 2017.
\newblock Abstractive document summarization with a graph-based attentional
  neural model.
\newblock In {\em Proceedings of the 55th Annual Meeting of the Association for
  Computational Linguistics (Volume 1: Long Papers)}, pages 1171--1181.

\bibitem[\protect\citename{Wang \bgroup et al.\egroup }2019]{wang2019self}
Hong Wang, Xin Wang, Wenhan Xiong, Mo~Yu, Xiaoxiao Guo, Shiyu Chang, and
  William~Yang Wang.
\newblock 2019.
\newblock Self-supervised learning for contextualized extractive summarization.
\newblock {\em arXiv preprint arXiv:1906.04466}.

\bibitem[\protect\citename{Wenbo \bgroup et al.\egroup }2019]{wenbo2019concept}
Wang Wenbo, Gao Yang, Huang Heyan, and Zhou Yuxiang.
\newblock 2019.
\newblock Concept pointer network for abstractive summarization.
\newblock {\em arXiv preprint arXiv:1910.08486}.

\bibitem[\protect\citename{Williams \bgroup et al.\egroup
  }2017]{williams2017broad}
Adina Williams, Nikita Nangia, and Samuel~R Bowman.
\newblock 2017.
\newblock A broad-coverage challenge corpus for sentence understanding through
  inference.
\newblock {\em arXiv preprint arXiv:1704.05426}.

\bibitem[\protect\citename{Wu and Hu}2018]{wu2018learning}
Yuxiang Wu and Baotian Hu.
\newblock 2018.
\newblock Learning to extract coherent summary via deep reinforcement learning.
\newblock In {\em Thirty-Second AAAI Conference on Artificial Intelligence}.

\bibitem[\protect\citename{Xiao and Carenini}2019]{xiao2019extractive}
Wen Xiao and Giuseppe Carenini.
\newblock 2019.
\newblock Extractive summarization of long documents by combining global and
  local context.
\newblock {\em arXiv preprint arXiv:1909.08089}.

\bibitem[\protect\citename{Xu and Durrett}2019]{xu2019neural}
Jiacheng Xu and Greg Durrett.
\newblock 2019.
\newblock Neural extractive text summarization with syntactic compression.
\newblock {\em arXiv preprint arXiv:1902.00863}.

\bibitem[\protect\citename{Yang \bgroup et al.\egroup
  }2019]{yang2019multilingual}
Yinfei Yang, Daniel Cer, Amin Ahmad, Mandy Guo, Jax Law, Noah Constant,
  Gustavo~Hernandez Abrego, Steve Yuan, Chris Tar, Yun-Hsuan Sung, et~al.
\newblock 2019.
\newblock Multilingual universal sentence encoder for semantic retrieval.
\newblock {\em arXiv preprint arXiv:1907.04307}.

\bibitem[\protect\citename{Zhang \bgroup et al.\egroup
  }2018]{zhang2018abstractiveness}
Fang-Fang Zhang, Jin-ge Yao, and Rui Yan.
\newblock 2018.
\newblock On the abstractiveness of neural document summarization.
\newblock In {\em Proceedings of the 2018 Conference on Empirical Methods in
  Natural Language Processing}, pages 785--790.

\bibitem[\protect\citename{Zhang \bgroup et al.\egroup
  }2019]{zhang2019bertscore}
Tianyi Zhang, Varsha Kishore, Felix Wu, Kilian~Q Weinberger, and Yoav Artzi.
\newblock 2019.
\newblock Bertscore: Evaluating text generation with bert.
\newblock {\em arXiv preprint arXiv:1904.09675}.

\bibitem[\protect\citename{Zhong \bgroup et al.\egroup
  }2019]{zhong2019searching}
Ming Zhong, Pengfei Liu, Danqing Wang, Xipeng Qiu, and Xuanjing Huang.
\newblock 2019.
\newblock Searching for effective neural extractive summarization: What works
  and what's next.
\newblock {\em arXiv preprint arXiv:1907.03491}.

\bibitem[\protect\citename{Zhou \bgroup et al.\egroup }2006]{zhou2006paraeval}
Liang Zhou, Chin-Yew Lin, Dragos~Stefan Munteanu, and Eduard Hovy.
\newblock 2006.
\newblock Paraeval: Using paraphrases to evaluate summaries automatically.
\newblock In {\em Proceedings of the main conference on Human Language
  Technology Conference of the North American Chapter of the Association of
  Computational Linguistics}, pages 447--454. Association for Computational
  Linguistics.

\bibitem[\protect\citename{Ziemski \bgroup et al.\egroup
  }2016]{ziemski2016united}
Micha{\l} Ziemski, Marcin Junczys-Dowmunt, and Bruno Pouliquen.
\newblock 2016.
\newblock The united nations parallel corpus v1. 0.
\newblock In {\em Proceedings of the Tenth International Conference on Language
  Resources and Evaluation (LREC'16)}, pages 3530--3534.

\end{thebibliography}
